\documentclass[conference]{IEEEtran}

\usepackage[switch]{lineno}  
\usepackage{enumitem}
\usepackage{amssymb}
\usepackage{mathtools}
\usepackage{caption}
\usepackage{subcaption}
\usepackage{amsmath}
\usepackage{amsfonts}
\usepackage{multirow}
\usepackage[ruled,vlined,linesnumbered]{algorithm2e}
\usepackage{algorithmic}
\usepackage{tabularx}
\usepackage{booktabs}
\usepackage{hyperref}
\usepackage{url}
\usepackage{fancyhdr}
\newcommand{\target}{\mathrm{tar}}
\newcommand{\train}{\mathrm{h}}
\newcommand{\test}{\mathrm{r}}

\newtheorem{definition}{Definition}
\newcommand{\local}{local }

\usepackage{color}

\definecolor{darkbrown}{rgb}{0.7,0.2,0.1}

\definecolor{orange}{rgb}{1,0.5,0}

\definecolor{darkgreen}{rgb}{0,0.5,0}

\definecolor{grey}{rgb}{0.7,0.7,0.7}

\begin{document}
\title{Space Meets Time: Local Spacetime Neural Network For Traffic Flow Forecasting}

\author{
\IEEEauthorblockN{
Song Yang,
Jiamou Liu, and
Kaiqi Zhao} 
\IEEEauthorblockA{School of Computer Science\\
The University of Auckland,
Auckland, New Zealand\\
syan382@aucklanduni.ac.nz, \{jiamou.liu, kaiqi.zhao\}@auckland.ac.nz}
}
\maketitle

\begin{abstract}
Traffic flow forecasting is a crucial task in urban computing. 
The challenge arises as traffic flows often exhibit intrinsic and latent spatio-temporal correlations that cannot be identified by extracting the spatial and temporal patterns of traffic data separately. 
We argue that such correlations are universal and play a pivotal role in traffic flow. We put forward {spacetime interval learning} as a paradigm to explicitly capture these correlations through a unified analysis of both spatial and temporal features. Unlike the state-of-the-art methods, which are restricted to a particular road network, we model the universal spatio-temporal correlations that are transferable from cities to cities. 
To this end, we propose a new spacetime interval learning framework that constructs a local-spacetime context of a traffic sensor comprising the data from its neighbors within close time points. Based on this idea, we introduce \local {spacetime neural network} (STNN), which employs novel spacetime convolution and attention mechanism to learn the universal spatio-temporal correlations. The proposed STNN captures local traffic patterns, which does not depend on a specific network structure. As a result, a trained STNN model can be applied on any unseen traffic networks.
We evaluate the proposed STNN on two public real-world traffic datasets and a simulated dataset on dynamic networks. The experiment results show that STNN not only improves prediction accuracy by 4\% over state-of-the-art methods, but is also effective in handling the case when the traffic network undergoes dynamic changes as well as the superior generalization capability.
\end{abstract}

\section{Introduction}\label{sec:intro}
With the increasing volume of traffic data and the growing impacts of data-driven technologies in modern transportation systems, the traffic flow forecasting problem is drawing increasing attention \cite{vlahogianni2014short}. Reliable and timely predictions of the traffic dynamics can assist transportation management, help alleviate traffic congestion, and enhance traffic efficiency. 

A traffic system is characterised by the changing flows at locations in a {\em road network}, i.e., nodes (sensors) interlinked by road segments, from which one may observe salient patterns: sudden bursts, drastic fluctuations, and periodic shifts. One can develop an understanding of these  patterns from two perspectives. On the outside, traffic is the accumulation of various extrinsic factors, from road layout, to geographic features of the environment, to traffic laws,  and erratic behaviours of drivers, all playing a role in shaping traffic conditions. On the inside, intrinsic principles are governing the flow of traffic, as vehicles -- particles in the road system -- maintain stable directional, speed and concentration features as they travel through space and time, giving rise to some {\em universal patterns} of the traffic flow. These universal patterns are what is behind well-established  traffic flow models, such as Wardrop's equilibria, and  Kerner's three-phase traffic theory, that describe traffic flow from a mathematical point of view \cite{treiber2013traffic}.  

We argue that accurate predictions of traffic flow -- based on macro-level traffic data -- rest upon a model's ability to grasp not only extrinsic features from the given data set, but also intrinsic principles that are universal to any traffic systems. At its core, traffic can be seen as a physical system \cite{yperman2005link} where localised changes cascade like waves along roads, to some other locations, some time later \cite{sugiyama2008traffic}. Therefore the key to capturing the intrinsic principles of traffic flow lies in extracting the latent correlations between locations' present states and surrounding locations' past. 

\begin{figure}[tb!]
	\centering
	\includegraphics[width=0.8\linewidth]{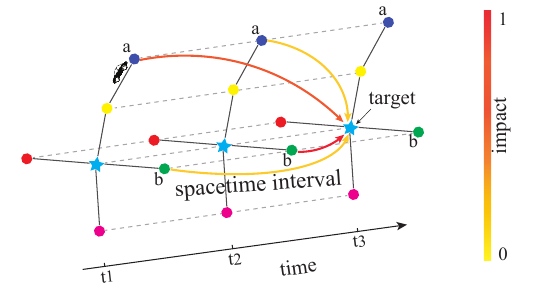}
	\caption{An illustration of spacetime interval. The figure shows three snapshots of the network. The target node's state at $t_3$ is strongly influenced by the state of $a$ at $t_1$ and the state of $b$ at $t_2$, but only weakly influenced by the state of $a$ at $t_2$ and the state of $b$ at $t_1$. }
	\label{fig:fig1}
\end{figure}

Recent breakthroughs in neural-based techniques, in particular those based on graph neural networks (GNNs)  \cite{yu2017spatio,yao2018deep,diao2019dynamic,zhang2020spatio}, represent a significant step towards representing non-linear traffic features by leveraging topological constraints while achieving better prediction results as compared to earlier statistical methods such as autoregressive models \cite{williams2003modeling}. However,  existing GNN-based traffic forecasting models have three common limitations: 
(1)  These models heavily rely on the graph structure which varies across different road networks. As such, they are restricted to a particular road network, rather than revealing intrinsic properties of traffic system;
(2) They apply computationally expensive feature aggregation operations (e.g., graph convolutions takes) on the entire network, and thus are not scalable to large road networks that contain hundreds of thousands of sensors. For instance, graph convolution network (GCN) scales quadratically w.r.t. the number of nodes in the network. 
(3) These models all consist of \emph{separate components} to extract features from the spatial and the temporal dimensions, respectively, before integrating them into the same feature map to derive a prediction. The extracted feature map represents ``aggregated'' correlations between locations of the network across time. This setup implicitly assumes that the correlations between locations stay uniform over time. However, two locations may have different correlations at different times: As indicated in Figure~\ref{fig:fig1}, against the current ($t_3$) target node, node $a$ has a stronger correlation at an earlier time (i.e., $t_1$) than a later time (i.e., $t_2$), while $b$ has a weaker correlation at $t_1$ than at $t_2$.

To break the above limitations in existing GNN-based models, we propose a new spatio-temporal correlation learning paradigm, called \emph{Spacetime Interval Learning}, which fuses the spatial dimensions and the temporal dimension into a single manifold, i.e., a {\em spacetime}, as coined in special relativity theory. The correlations that we aim to capture reflects a form of ``distance'' between two traffic events in spacetime, where the distance is referred to as {\em interval} in spacetime terminologies \cite{rowe2013geometrical}. Specifically, the proposed spacetime interval learning paradigm extracts the traffic data of nearby sensors within a fixed time window, regarded as the \emph{local-spacetime context} of each target location. The local-spacetime context is analogous to the ``receptive field'' of sensors to ignore the sensors that do not contribute much to the final prediction.
Then, the spacetime interval learning paradigm learns to correlate a node $x$ at a given time $t$ with another node $y$ at a different time $t'\geq t$ in the local-spacetime context. 

The proposed learning paradigm addresses the limitations of the GNN-based models as follows. Firstly, it models the spatio-temporal correlations within the local-spacetime context of each target location. In this way, the model we build is independent of the graph structure, and thus is {\em universal} for traffic flows at large, as opposed to for any specific network. Secondly, our method focuses on the local-spacetime, such that it changes the traffic prediction from \emph{network-level} to \emph{node-level}. That is, our model can be predict the traffic for multiple locations of interest in parallel on different machines. 
Thirdly, unlike existing methods, we fuse the spatial and the temporal dimensions into a single manifold to capture the varying spatial correlations between locations across time.

Under the spacetime interval learning paradigm, we propose an implementation called Spacetime Neural Network (STNN), which combines novel {\em spacetime attention blocks} and {\em spacetime convolutional blocks}. The former highlights the interval between events, while the latter aggregates the learned features from spatial, temporal, and spatio-temporal aspects.
The main advantage of the model are as follows. The code is available on \url{https://github.com/songyangco/STNN}.

\begin{enumerate}[leftmargin=*]
\item {\em Accuracy}: The learned spacetime intervals explicitly capture the correlations between locations across different time instances, making the learned features highly informative in traffic prediction. As validated through experiments on two real-world datasets, METR-LA and PeMS-Bay, when trained on data from a single network, our model outperforms existing methods, with, e.g., 4\% improvement over current state-of-the-art complicated models for 15-min predictions. See Section~\ref{sec:realworld}.

\item {\em Transferrability} amounts to a main advantage of our universal model. For this, we train a single STNN model on one dataset and use it  without fine-tuning on unseen traffic datasets. In all cases, the results achieve competitive, if not better, performance compared to state-of-the-art approaches, validating the applicability of our universal model. See Section~\ref{sec:realworld}.

\item {\em Ability to handle dynamic networks} is another natural consequence of converting raw input to a local-spacetime for every target node. 
This refers to, e.g., the  network undergoes structural changes such as the blocking or addition of roads. We perform an experiment on a synthetic dynamic network. Our model is much better than baselines. See Section~\ref{sec:simulated}.

\item {\em Scalability}: Previous GNN-based models all assume a small fixed network whose size is limited by computational capability. By performing prediction on node-level, we break the limit on the network size. The complexity of our model is independent from the network size, making our model applicable to networks of arbitrary size. See Section~\ref{sec:complexity}.
\end{enumerate}

\section{Related Works}\label{sec:related}
Studies on traffic forecasting through spatio-temporal data analysis can roughly be divided into four types.

\textbf{Statistical methods}. 
Early studies in the 2000s use various statistical methods, such as historical average (HA), autoregressive integrated moving average (ARIMA) \cite{williams2003modeling}, and vector autoregression (VAR) \cite{zivot2006vector} for the traffic forecasting problem. However, these approaches assume the input time series to be stationary and univariate. 
This assumption may not hold in a complex traffic system. Thus, statistical methods often generate inaccurate predictions. 

\textbf{Machine learning methods}.
To account for more complex traffic data, various classical machine learning methods, such as kNN \cite{van2012short}, SVM \cite{zhang2011seasonal}, SVR \cite{chen2015forecasting}, are employed in the early 2010s. One shortcoming of these methods is that they fail to capture the highly non-linear spatial and temporal patterns present in the data.

\textbf{Deep learning methods}.
After 2015, capitalising on the success of deep learning \cite{lecun2015deep}, several deep learning models are proposed to capture the underlying patterns of traffic data. Recurrent neural networks (RNNs) are adopted to analyse time series patterns \cite{laptev2017time,cui2018deep}, while convolutional neural networks (CNNs) are used to capture spatial dependencies by treating a traffic network as a grid \cite{zhang2016dnn,zhang2017deep}. However, these methods omit the road network, which is an essential spatial constraint on how traffic moves in the spatial dimension. This limitation inhibits traditional deep learning methods from accurate traffic forecasting.

\textbf{Graph neural networks}.
More recently, graph-based deep learning models are increasingly used for spatio-temporal data analysis. Yu et al. \cite{yu2017spatio} propose a spatio-temporal graph convolutional network based on spectral graph theory to extract features from the road network and the historical time series data. In particular, the spatial patterns are captured by graph convolutional layers while the temporal patterns are extracted by gated convolutional layers. Li et al. \cite{li2017diffusion} model the traffic flow as a diffusion process and capture the spatial dependencies by random walks on a graph followed by an RNN to extract temporal patterns. Yao et al. \cite{yao2018deep} use graph embeddings to capture the road network information and integrate Long-Short Term Memory (LSTM) and CNN with the graph embeddings for traffic prediction. Attention mechanism has also been adopted to learn spatial and temporal patterns with improved performance~\cite{zhang2018gaan,guo2019attention}. However, all of these methods fall into the same paradigm that separate layers are applied to extract spatial patterns and temporal patterns respectively. We summarize current methods under this paradigm in Table~\ref{tab:existing_methods}. Particularly, CNN, Graph Convolutional Network (GCN), or spatial attention are often used to learn spatial patterns. Gated Recurrent Unit (GRU), LSTM, gated temporal convolution (gated TCN), or temporal attention are used to learn temporal patterns.

\begin{table}
\small
  \caption{Existing methods fall into the same paradigm}
  \label{tab:existing_methods}
  \begin{tabular}{ccc}
    \toprule
    {Studies} & {Spatial layers} & {Temporal layers}\\
    \midrule
    DCRNN (2017) \cite{li2017diffusion} & Diffusion Conv &  GRU \\
    STGCN (2017) \cite{yu2017spatio} & GCN & Gated TCN\\
    DMVST-Net (2018) \cite{yao2018deep} & CNN & LSTM \\
    STDN (2019) \cite{yao2019revisiting} & CNN & LSTM + Attn \\
    GraphWaveNet (2019) \cite{wu2019graph} & GCN & Gated TCN \\
    DSTGCN (2019) \cite{diao2019dynamic} & GCN & TCN \\
    ASTGCN (2019) \cite{guo2019attention} & GCN + Attn & TCN + Attn\\
    SLCNN (2020) \cite{zhang2020spatio} & GCN & Gated TCN \\
    GMAN (2020) \cite{zheng2020gman} & Spatial Attn & Temporal Attn \\
    MTGCN (2020) \cite{wu2020connecting} & GCN & Gated TCN \\
    \hline
    \hline
    STNN (ours) & \multicolumn{2}{c}{ST Conv + Attn} \\
  \bottomrule
\end{tabular}
\end{table}

Unlike current methods, this work proposes a novel learning paradigm. Instead of using separate components or layers to learn spatial and temporal patterns, we design a novel spacetime module to learn the local spatio-temporal correlations, and capture both spatial and temporal patterns at the same time. Our model does not fall into any existing categories. Specially, instead of exploiting GNN to learn spatial patterns, we fuse spatial features with temporal features together and learned by a unified layer.

\section{Problem Formulation} 
We summarise the list of important notations in Table~\ref{tab:notations}.

\begin{table}[thbp]
\caption{Table of Notations}
\label{tab:notations}
\small
\begin{tabular}{l|l}
\toprule
Notation &  Description \\
\midrule
$V^{(t)}$ & sensor set at time $t$   \\ \hline
$V$ & union sensor set of all time steps  \\ \hline
$Q^{(t)}$ & distance matrix at time $t$\\ \hline
$\mathcal{Q}$ & stack $Q^{(t)}$ along time axis\\ \hline
$\mathbf{x}_i^{(t)}$ & a vector of features recorded by $i$-th sensor at time $t$ \\ \hline
$\mathbf{X}^{(t)}$ & a matrix of features of all sensors at time $t$ \\ \hline
$\mathcal{X}$ & stack $\mathbf{X}^{(t)}$ along time axis\\ \hline
$V_{\text{tar}}$ & a set of target nodes to predict\\ \hline
$v_p$ & a target node   \\ \hline
$V_p$ & a set of neighboring nodes that close to $v_p$ \\ \hline
$F$ & number of features captured by each sensor station \\ \hline
${y}_i^{(t)}$ & future traffic measure of $i$-th sensor at time $t$  \\ \hline
$\mathbf{y}_i$ & future traffic measures of $i$-th sensor  \\ \hline
$\mathcal{Y}$ & future traffic measures of all nodes in $V_{\text{tar}}$  \\ \hline
$\mathbf{A}^{(t)}$ & normalized connectivity matrix at time $t$  \\ \hline
$\mathbf{D}_p^{(t)}$ & a local-spacetime snapshot for $v_p$ at time $t$  \\ \hline
$\mathcal{D}_p$ & a local-spacetime for $v_p$ \\ 
\bottomrule
\end{tabular}
\end{table}

Traffic condition like vehicle speed is often recorded by the road-side sensor station $v$ along with other auxiliary information including time, sensor location, etc. Aggregating $N$ sensors together forms a sensor set $V \in \mathbb{R}^{N}$ which monitor the traffic flow of a certain area.

At each time step $t$, the $i$-th sensor station $v_i$ records local traffic measures such as vehicle average speed in a fixed time duration and supplementary data. These measurements form a feature vector $\mathbf{x}^{(t)}_i\in \mathbb{R}^F$. In particular, we have two features in our input data: average speed, timestamp. Collecting the feature vectors of all sensor stations together, we obtain the overall observation of the traffic network at time step $t$ as $\mathbf{X}^{(t)}=\left(\mathbf{x}^{(t)}_1,\mathbf{x}^{(t)}_2,\ldots,\mathbf{x}^{(t)}_i\right) \in \mathbb{R}^{N \times F}$. A traffic flow dataset contains the measures of a sequence of $T_h$ time steps, where $T_h>0$, and is thus presented as a {\em sensor feature tensor} $\mathcal{X}\coloneqq \left(\mathbf{X}^{(1)},\mathbf{X}^{(2)},\ldots,\mathbf{X}^{(T_\train)}\right) \in \mathbb{R}^{N\times F \times T_\train}$. 

Based on the location of sensor stations, we can get the travel distance on road network between any two sensors. Consequently, we built a matrix $\mathbf{Q}$ to reflect the distances of all sensor pairs. One can think $\mathbf{Q}$ as the weighted adjacent matrix of a directed complete graph. The edge weight is affected by the actual distance and complete graph is because we can always calculate the distance between two sensors. Furthermore, we do not require $\mathbf{Q}$ remain unchanged all the time. On the contrary, the travel distance between two sensors might be variable because of events like traffic accidents or road constructions which alter the underlying topology of road network, leading to a further travel distance. Thus, to better incorporate the network dynamics in our model, we use $\mathbf{Q}^{(t)}$ to just indicate the spatial context at the particular time step $t$. $\mathcal{Q}\coloneqq \left(\mathbf{Q}^{(1)},\mathbf{Q}^{(2)},\dots,\mathbf{Q}^{(T_\train)} \right) \in \mathbb{R}^{N\times N \times T_\train}$ reflects the underlying road network structure dynamics over $T_\train$ time steps, where $T_\train>0$. For a static network, $\mathcal{Q}$ reduces to $\textbf{Q}$.

\textbf{Traffic flow forecasting}: Given a sensor feature tensor $\mathcal{X}$ and a spatial context tensor $\mathcal{Q}$ over past $T_\train$ time steps, as well as a set of nominated {\em target nodes} $V_\target=\{v_1,v_2,\ldots,v_M\}\subseteq V$. The traffic flow forecasting problem aims to predict the traffic flow of the next $T_\test$ time steps for every target node in $V_\target$. The desired output is thus $\mathcal{Y}=(\mathbf{y}_1,\mathbf{y}_2,\dots,\mathbf{y}_M) \in \mathbb{R}^{M \times T_\test}$ where $\mathbf{y}_p =  (y_p^{T_\train+1},y_p^{T_\train+2},\dots,y_p^{T_\train+T_\test}) \in \mathbb{R}^{T_\test}$ denotes the traffic conditions of the next $T_\test$ time steps at node $v_p\in V_\target$.

\section{Spacetime Interval Learning}
{\em Spacetime interval learning} aims to discover the influence between traffic events in a single {\em spacetime manifold} instead of modeling the spatial and temporal dimensions separately. 
We define traffic events and spacetime as follows:

\begin{definition}[\textbf{Traffic Event}]
Given a traffic measurement $s$ (e.g., speed) observed at sensor $v_i$ and time $t$, a traffic event is a tuple consists of the measurement, time, and location, namely, $(s,t, v_i)$. 
\end{definition}

A traffic event, defined analogously to the notion of events in physics \cite{carroll2019spacetime}, embodies both  spatial and temporal information of a traffic measurement by a sensor station. 

\begin{definition}[\textbf{Spacetime}]
Given a subset of sensors $V^\prime \subset V$, the sensor features $\mathbf{X}^\prime$, the related spatial context $\mathbf{Q}^\prime$, and the $T$ time steps historical data, the spacetime $\mathcal{D}$ is a manifold consists of all traffic events produced by $\{\mathbf{X}^{\prime(1)},\mathbf{X}^{\prime(2)},\cdots,\mathbf{X}^{\prime(T)}\}$ and $\{\mathbf{Q}^{\prime(1)},\mathbf{Q}^{\prime(2)},\cdots,\mathbf{Q}^{\prime(T)}\}$.
\end{definition}

A spacetime manifold can be organzied as a 3D structure where first dimension corresponding to the time, the second dimension corresponding to the space, and the third is the observed traffic measurement. The time dimension is easy to understand but the space data needs more work. In order to squeeze the spatial information into a single dimension, we transfer the sensor coordinates to the travel distance with the anchor/target sensor. As a result, we construct a local-spacetime around a target sensor and use that to predict the future traffic flow of that particular sensor.

\begin{definition}[\textbf{Local-spacetime}]
The local-spacetime $\mathcal{D}_p$ for sensor $v_{p}$ is a subset of $\mathcal{D}$, which only contains the traffic events occur at nearby location of $v_{p}$ and recent time steps. Furthermore, the sensor location in all traffic events from $\mathcal{D}_p$ is replaced by its travel distance with $v_{p}$.
\end{definition}

The {\em spacetime interval} between two traffic events denotes the extent to which one event influences the other; a smaller interval means a closer association between two traffic events. In a local-spacetime, we only care about the intervals between traffic events of the target sensor with other sensors.
\begin{definition}[\textbf{Spacetime Interval}]
Spacetime interval is the quantified influence of a traffic event imposed on another traffic event regarding to the traffic measurement.
\end{definition}

A crucial step in the proposed learning paradigm is to build the local-spacetime of a target node $v_{p}$, which is then used for spacetime interval learning. In this way, the trained model not only captures the spacetime interval explicitly, but also be able to generalize learned patterns to other local-spacetimes in the same road network or event different city. We now give details of how to construct a local-spacetime for an arbitrary node $v_p\in V_\target$ where $V_\target$ is a set of nodes we want to predict. The pseudocode is outlined in Algorithm~\ref{alg:algorithm1}.

\begin{algorithm}[!htbp]
\small
\selectfont
	\caption{Local-spacetime construction}
	\label{alg:algorithm1}
	
	\SetAlgoLined
	\SetKwInOut{Parameter}{Parameter}
	\SetKwInOut{Input}{Input}
	\SetKwInOut{Output}{Output}
	
	\KwIn{
	$\mathcal{X}=\left(\mathbf{X}^{(1)},\mathbf{X}^{(2)},\dots,\mathbf{X}^{(T_\train)}\right) \in \mathbb{R}^{N\times F \times T_\train}$
	\newline
	$\mathcal{Q}=\left(\mathbf{Q}^{(1)},\mathbf{Q}^{(2)},\dots,\mathbf{Q}^{(T_\train)} \right) \in \mathbb{R}^{N\times N \times T_\train}$ \newline		
	    Target nodes: $V_\target=\{v_1,v_2,\ldots,v_M\}$ 
	    }
	\Parameter{$\epsilon>0$, $\theta > 0$ }
	\KwOut{Local-spacetime list $\{\mathcal{D}_1,\mathcal{D}_2,\dots,\mathcal{D}_M\}$}
	Set $\mathbf{A}^{(t)} \leftarrow \exp\left(-{{\left(\mathbf{Q}^{(t)}\right)^2}}/{{\theta^2}}\right)$ $\forall t=1,\ldots,T_\train$\;
	\For{$v_p \in V_\target$}{
		\For{$t \leq T_\train$}{
			\If{$\mathbf{A}^{(t)}(i,p) > \epsilon$ or $\mathbf{A}^{(t)}(p,i) > \epsilon$}{
				Set $V_p\leftarrow V_p\cup \{v_i\}$;%
			}
		}
		\For{$t \leq T_\train$ }{
		    \For{$v_i \in V_p$}{
		        Set $\mathbf{D}_{p}^{(t)}(i) \leftarrow   \left[\mathbf{x}^{(t)}_i;\mathbf{A}^{(t)}({i,p})\right]\in \mathbb{R}^{F+1}$;
		    }
		}
		$\mathcal{D}_p\leftarrow \left(\mathbf{D}_p^{(1)},\ldots,\mathbf{D}_p^{(T_\train)} \right)  \in \mathbb{R}^{|V_p| \times (F+1) \times T_\train}$\;
	}
\end{algorithm}
Given a time step $t$, we define a connectivity matrix $\mathbf{A}^{(t)}\in {V^{(t)}}^2\to [0,1]$ by applying the Gaussian kernel on the distance matrix $\mathbf{Q}^{(t)}$ to convert travel distance to a weight reflect the connectivity of two sensors. Longer distance corresponds to smaller weight, namely,
\begin{equation}\label{eqn:A}
  \mathbf{A}^{(t)}(i,j) \coloneqq \exp\left(-\frac{{{\mathbf{Q}^{(t)}(i,j)}^2}}{{\theta^2}}\right)\in [0,1]
\end{equation}
where $i$, $j$ are any two nodes in $V^{(t)}$ and $\theta$ is a hyper-parameter. Empirically, we set $\theta$ as the standard deviation among all $\mathbf{Q}^{(t)}(i,j)$.  Intuitively, $\mathbf{A}^{(t)}(i,j)$ is a normalized value that expresses how easy to travel from $v_i$ to $v_j$ in the network at time step $t$. Note that Equation~\eqref{eqn:A} guarantees $\mathbf{A}^{(t)}(i,i)=1$. The superscript $t$ used to depict the variational travel distance caused by the underlying networks structure change.

Using $\mathbf{A}^{(t)}$, we extract a set of nodes, denoted as $V_p$, that could benefit the future traffic flow prediction for the target node $v_p$ as 
\begin{equation}
\label{equ:V_p}
 V_p:=\left\{v_i\bigg| \max\{\mathbf{A}^{(t)}(i,p),\mathbf{A}^{(t)}(p,i)\mid 1\leq t\leq T_\train\} > \epsilon \right\}
\end{equation}
where $\epsilon$ is a pre-determined threshold parameter that indicates  how close a node should be from (or to) $v_p$ to be considered relevant to the traffic flow at $v_p$. Note that $\mathbf{A}^{(t)}(i,p)$ can be different from $\mathbf{A}^{(t)}(p,i)$ because $\mathbf{Q}^{(t)}$ is not symmetric. $\epsilon$ control the trade-off between computational cost and prediction accuracy. Moreover, to have the fixed shape input data for training, we require the size of $V_p$ to be $\alpha$, namely, $|V_p| = \alpha$. This is done by keeping the nearest $\alpha$ neighbors only if $|V_p| > \alpha$ or add dummy nodes if  $|V_p| < \alpha$. Dummy nodes have no connections with other nodes, and their features are all zero. As a results, nodes in $V_p$ are sorted based on the travel distance to $v_p$ in ascending order.

We now construct a local-spacetime $\mathcal{D}_p$ of $v_p$. For $t\in \{1,\ldots,T_\train\}$, each row of the matrix $\mathbf{D}_p^{(t)} \in \mathbb{R}^{|V_p|\times (F+1)}$ is defined as:
\begin{equation}
\mathbf{D}_p^{(t)}(v_i)\leftarrow \left[\mathbf{x}^{(t)}_i;\mathbf{A}^{(t)}(i,p)\right] \text{, } v_i\in V_p
\end{equation}
where $\mathbf{x}^{(t)}_i$ is the traffic measurement recorded at node $v_i$ and time step $t$, and $[\cdot;\cdot]$ denotes concatenation. 
The matrix $\mathbf{D}_p^{(t)}$ can be regarded as a snapshot of local-spacetime $\mathcal{D}_p$. It encodes spatial relationship between $v_p$ and those nodes in $V_p$, as well as traffic measurements of all these nodes at time step $t$. Finally, define the local-spacetime $\mathcal{D}_p$ by
\begin{equation}
\mathcal{D}_p \coloneqq \left(\mathbf{D}_p^{(1)},\mathbf{D}_p^{(2)},\dots,\mathbf{D}_p^{(T_\train)}\right)  \in \mathbb{R}^{|V_p| \times (F+1) \times T_\train}
\end{equation}
Since $\mathcal{D}_p$ contains all the information we need to train a predictive model of the future traffic flow at $v_p$, the learning process is independent on the size of the entire network, thus resolving the scalability issue. Moreover, the incorporation of the network information at all time steps $t\in \{1,\ldots,T_\train\}$ makes the model capable of handling dynamic network topology. 

In summary, the spacetime interval learning framework reduces the traffic flow forecasting problem to learning a universal function $g(\cdot)$ that maps $\mathcal{D}_p$  to the future traffic flow of $v_p$ and $p \in \{1,\dots,n\}$:
\begin{equation}
\label{equ:main_equ}
\left(\mathbf{D}_p^{(1)},\mathbf{D}_p^{(2)},\dots,\mathbf{D}_p^{(T_\train)} \right)  \stackrel{g(\cdot)}{\longrightarrow} \left(y_p^{T_\train+1},\dots,y_p^{T_\train+T_\test}\right)
\end{equation}
In next section, we propose a novel model as a realization of the mapping function $g(\cdot)$, namely, the spacetime neural network.

\begin{figure*}
	\centering
	\includegraphics[width=\linewidth]{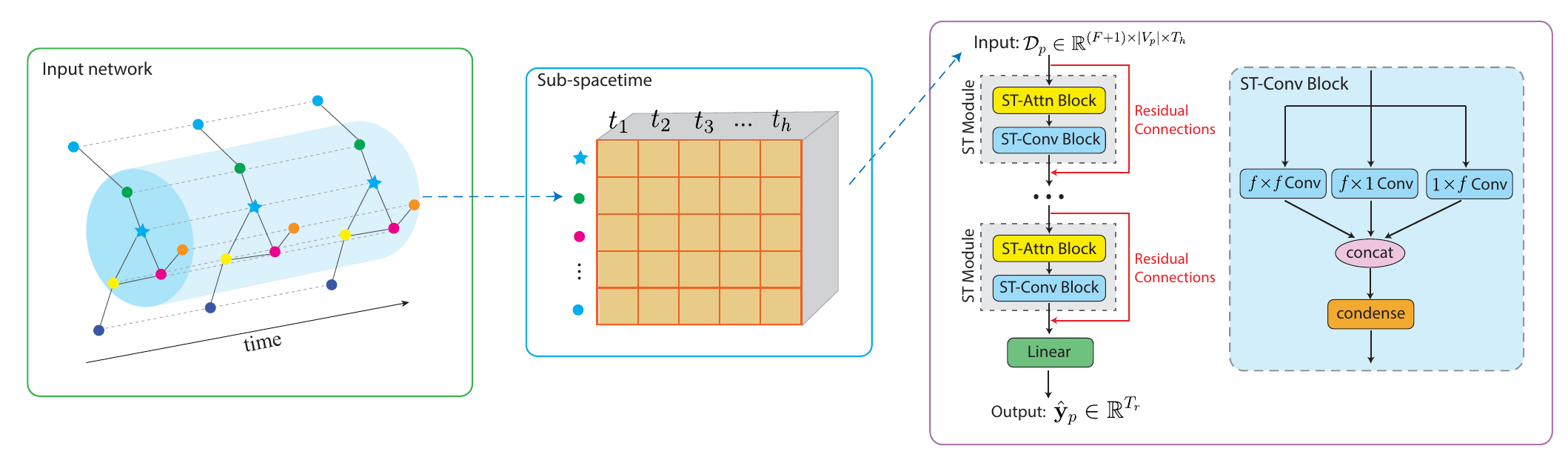}
	\caption{The architecture of STNN with an example local-spacetime constructed from the input data. STNN consists of $k$ spacetime modules (ST-Modules) and a fully-connected output layer. Each ST-Module contains a spacetime attention block (ST-Attn block) and a spacetime convolution block (ST-Conv block). The ST-Attn block uses self-attention mechanism to spotlight the most contributive traffic events. In each ST-Conv block, three different convolution kernels are employed to aggregate the spatio-temporal correlations in different perspectives. Then, the extracted features are stacked, and condensed by the $1\times 1$ convolution. 
	}
	\label{fig:model-structure}
	\vspace{-0.2cm}
\end{figure*}

\section{Spacetime Neural Network}
The core task of spacetime interval learning is to adaptively learn the influence between traffic events and use them to predict the future traffic flow. To this end, we propose {\em SpaceTime Neural Network (STNN)}, which is an end-to-end deep learning model that realizes the function $g(\cdot)$ in Equation~\ref{equ:main_equ}.

\textbf{Design Principles.}
To quantify the influence of traffic events in the local-spacetime to the prediction of future traffic flow on the target sensor, a natural idea is to learn how each single traffic event influences the prediction. However, we observe that traffic events in local-spacetime are not independent and they may interfere each other. For example, congestion in an arterial road may increase the traffic in a surrounding region and then diffuse to further road segments. As such, we design our model with two main principles:  (1) the model should be able to capture the pair-wise influence, and (2) it needs to be aware of the conditions of nearby regions, capturing the many-to-one influence. 

\textbf{Model Architecture.}
Following the above two principles, we propose a novel spacetime module, on top of which we build the proposed STNN model.
A spacetime module comprises a {\em Spacetime Attention Block} (Section~\ref{sec:spacetime_attn}) to capture the pair-wise influence, and a {\em Spacetime Convolution Block} (Section~\ref{sec:spacetime_conv}) to capture the many-to-one influence.

The overall architecture of STNN is shown in Figure~\ref{fig:model-structure}. The input is a local-spacetime of the target (central) node represented by the blue cylinder. STNN employs several spacetime modules to predict the future traffic of the target node.

More formally, convert the raw data to the local-spacetime $\mathcal{D}_p$ and permute $\mathcal{D}_p$ to the channel-first format, i.e., $\mathcal{D}_p \in \mathbb{R}^{(F+1)\times N \times T_\train}$. Then the input $\mathcal{D}_p$ is transformed using $1\times 1$ convolution to map traffic events to a high-dimensional feature space. Next, we stack $k$ spacetime modules where $k$ controls the trade-off between model complexity and performance. A smaller $k$ leads to a model with fewer parameters and unlikely to overfits the data. A larger $k$ gives better performance but may jeopardize the generalization ability. Residual connections are also applied for each module. Last, we use a fully connected layer with linear activation to present the prediction results $\left(\hat{y}_p^{T_\train+1},\dots,\hat{y}_p^{T_\train+T_\test}\right)$.

\subsection{Spacetime Attention Block}\label{sec:spacetime_attn}
Inspired by self-attention mechanism \cite{zhang2019self}, we propose the {\em spacetime attention} block to automatically discover the pairwise influence between traffic events. Specifically, for the attention block at the $l$~th layer, the input is a 3D tensor which is the output of previous convolutional layer or the initial input data, i.e., $\mathcal{D}_p^{[l-1]} \in \mathbb{R}^{C^{[l-1]} \times |V_p| \times T_\train}$, where $C^{[l-1]}$ denotes the number of channels for feature maps output by the previous layer. Note that $\mathcal{D}_p^{[0]}=\mathcal{D}_p$. Let $U=|V_p| \times T_\train$ and we calculate the attention map as
\begin{equation}
\boldsymbol{S} = (\boldsymbol{W}_{q} \mathcal{D}_p^{[l-1]})^T \boldsymbol{W}_{k} \mathcal{D}_p^{[l-1]} \in \mathbb{R}^{U \times U}
\end{equation}
\begin{equation}
\boldsymbol{S}^{\prime}_{i,j}=\frac{\exp \left(\boldsymbol{S}_{i, j} 
\right)}{\sum_{j=1}^{U} \exp \left(\boldsymbol{S}_{i, j} 
\right)}
\end{equation}
where $\boldsymbol{W}_{q} \in \mathbb{R}^{C^{[l]} \times C^{[l-1]}}$ and  $\boldsymbol{W}_{k} \in \mathbb{R}^{C^{[l]} \times C^{[l-1]}}$ are learnable weight matrices that project each traffic event into feature spaces corresponding to ``query'' and ``key'', respectively. As such, the dot product of the ``query'' and ``key'', i.e., $\boldsymbol{S}_{i, j}$ indicates the relevance between the $i$-th and $j$-th traffic events. The softmax function is used to guarantee the attention weights sum to one. Then, we apply another learnable weight matrix $\boldsymbol{W}_{v} \in \mathbb{R}^{C^{[l]} \times C^{[l-1]}}$ to project input data into an output feature space and use the attention matrix to weight the contribution from different traffic events:
\begin{equation}
\mathcal{D}_p^{[l]} = \boldsymbol{W}_{v} \mathcal{D}_p^{[l-1]} \boldsymbol{S}^{{\prime}^T} \in \mathbb{R}^{C^{[l]}\times |V_p| \times T_\train}
\end{equation}
The attention block can dynamically adjust the impact of different traffic events to target events w.r.t. the features from the previous layer.

\subsection{Spacetime Convolution Block}\label{sec:spacetime_conv}
\begin{figure}
	\centering
	\includegraphics[width=0.44\linewidth]{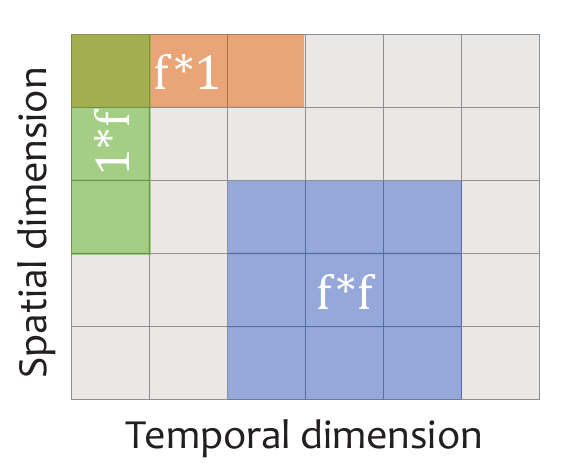}
	\caption{Spatial kernel, temporal kernel and spacetime kernel on the sub-spacetime}
	\label{fig:st_conv}
	\vspace{-0.2cm}
\end{figure}

After the attention layer, we design a {\em spacetime convolution block} that contains three kernels to capture the many-to-one influence from three different perspectives corresponding to space, time, and spacetime, respectively. The {\em spacetime kernel} is the main perspective for uncovering the spacetime correlations from a local local-spacetime to the target events. The {\em temporal kernel} finds correlations between traffic events of the same location along time, and the {\em spatial kernel} captures the influence from nearby locations on the same time step. The motivation for adopting the temporal kernel is that each sensor may have unique periodic temporal traffic patterns (peak/off-peak, weekday/weekend, etc.), which may be undervalued from the spacetime perspective. Similarly, we use the spatial kernel to keep the underlying geo-spatial influence from neighbors which is invariant to the time. Figure~\ref{fig:st_conv} demonstrates these three kernels on the spacetime manifold. 
Each spacetime convolution block takes the output of former attention block as the input, e.g., $\mathcal{D}_p^{[l]} \in \mathbb{R}^{C^{[l]} \times |V_p| \times T_\train }$. The output $\mathcal{D}_p^{[l+1]} \in \mathbb{R}^{C^{[l+1]} \times |V_p| \times T_\train }$ is computed by
\begin{equation}
H = \text{LeakyReLU}\left[\Theta^{[l+1]}_{st} * \mathcal{D}^{[l]}_{p};  \Theta^{[l+1]}_{t} * \mathcal{D}^{[l]}_{p}; \Theta^{[l+1]}_{s} * \mathcal{D}^{[l]}_{p}\right]
\end{equation}
\begin{equation}
\mathcal{D}_p^{[l+1]} = \text{LeakyReLU}\left(\Theta^{[l+1]}_o * H\right),
\end{equation}
where $\Theta^{[l+1]}_{st}$, $\Theta^{[l+1]}_{t}$, and $\Theta^{[l+1]}_{s}$ are the $f\times f$ spacetime kernel, the $f\times 1$ temporal kernel and the $1\times f$ spatial kernel, respectively; $\text{LeakyReLU}(\cdot)$ denotes the leaky rectified linear units function; and $*$ represents the convolution operation. For the above three convolution kernels, we take $f=3$ in the experiments. Additionally, padding is used to make sure the output has the same size as the input. Last, we concatenate the output of the three kernels and use an $1\times 1$ convolution, $\Theta^{[l+1]}_o$, to condense features and restrict the number of channels. 

At last, a fully connected layer is adopted to predict the future traffic flow from learned features. 

\section{Experiments}
To evaluate the performance of the proposed model, we compare STNN with the state-of-the-art traffic prediction models on two public traffic datasets in Section \ref{sec:realworld}. In addition, a simulated dynamic traffic network is used to demonstrate our model's capability of handling dynamic graphs in Section~\ref{sec:simulated}. We further investigate the learned spacetime interval via a case study (Section~\ref{sec:case_study}) and the impacts of different components of STNN (Section~\ref{sec:ablation}). 

\subsection{Experimental Setup}
\subsubsection{Real-world Networks}
Two public real-world datasets METR-LA and PeMS-Bay are used for evaluation. METR-LA were collected from the loop detectors in the highway of Los Angeles County \cite{jagadish2014big}, and PeMS-Bay were collected from California  Caltrans Performance Measurement System (PeMS) \cite{chen2001freeway}. These datasets were released by \cite{li2017diffusion} and have been widely used to evaluate traffic prediction models. METR-LA records four months of traffic data in early 2012 in Los Angeles County with 207 sensors. PeMS-Bay captures six months of traffic data in early 2017 in the Bay Area with 325 sensors. The speed reading of sensors in METR-LA and PeMS-Bay are aggregated to 5-minute windows, resulting in 288 data points per day. Missing values are filled with linear interpolation. Statistics of the datasets are summarized in Table \ref{tbl:dataset-statistics}.

\begin{table}[!htbp]
	\centering
	\caption{Statistics of datasets}
	\begin{tabular}{lllll}
		\toprule
		Data      & Nodes &Time steps & Traffic events& Dynamics \\ 
		\midrule
		METR-LA   & 207 & 34,272    &7,094,304 & No         \\
		PeMS-Bay  & 325 & 52,116    & 16,937,700 & No       \\
		Simulated & 84 & 2,000     & 16,800& Yes            \\
		\bottomrule
	\end{tabular}
	\label{tbl:dataset-statistics}
\end{table}
\subsubsection{Simulated Network}
\begin{figure}[!htbp]
	\centering
	\includegraphics[width=0.75\linewidth]{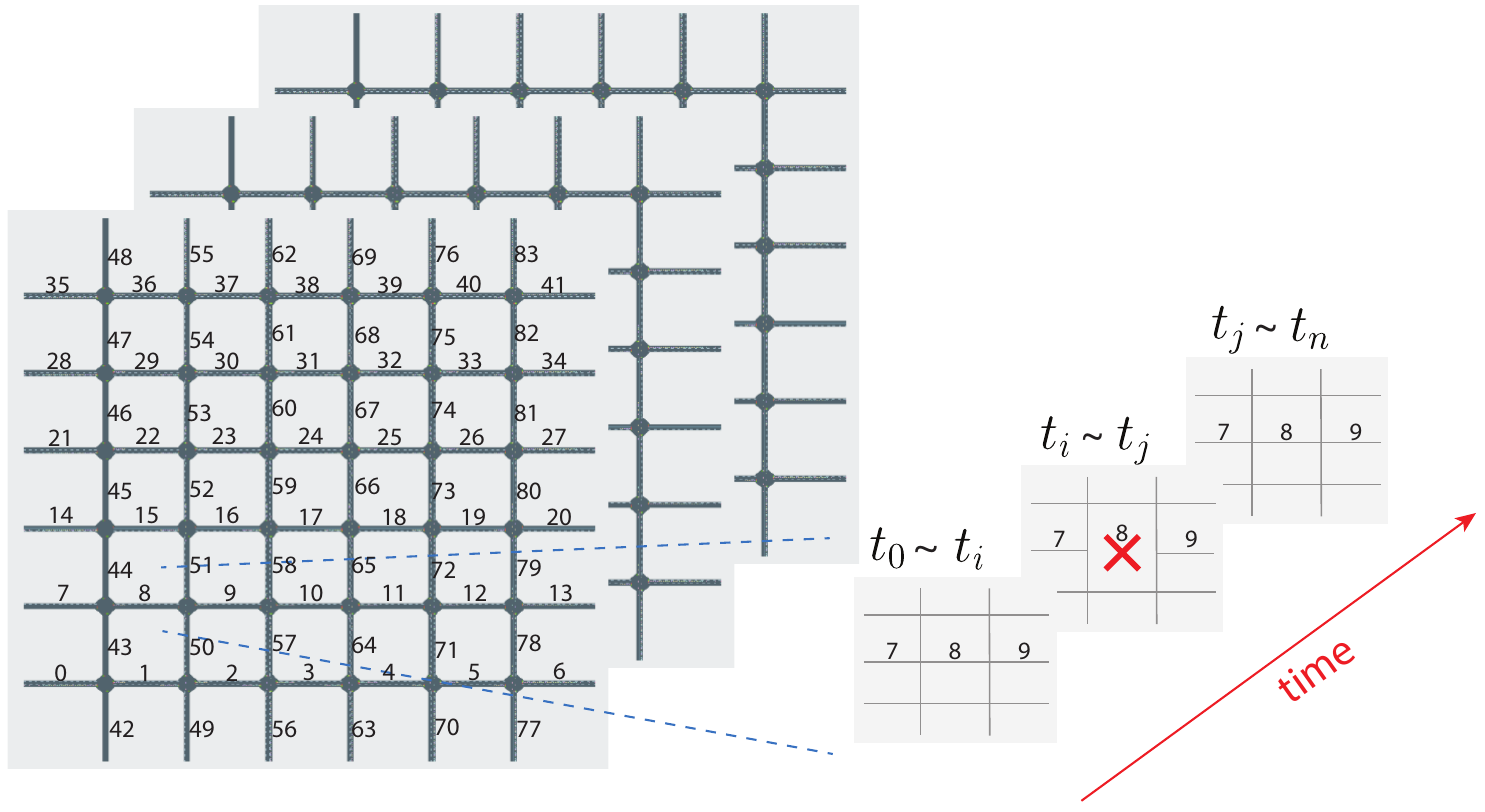}
	\caption{Simulated road network illustration}
	\label{fig:dynamic-network-demo}
\end{figure}

Apart from the two public datasets with static networks, we simulate a traffic network to demonstrate our model's ability to handle dynamic networks. We synthesize the dynamic traffic data using CityFlow \cite{zhang2019cityflow} which is a multi-agent reinforcement learning tool for city traffic scenarios. It is worthy to note that the goal of this experiment is not to show how the simulated traffic network resembles reality but to evaluate how well our model predicts traffic flow as the network topology changes. The simulated dataset contains 84 nodes and 2000 time steps. Each node denotes a sensor station located in the middle of a road segment as shown in Figure~\ref{fig:dynamic-network-demo}. Each sensor station records the total number of vehicles passing it in the fixed time interval instead of the average speed. Dynamics are represented as road closures to simulate traffic accidents or road constructions. For example, road segment 8 in Figure \ref{fig:dynamic-network-demo} is closed from 400 to 600 and 1500 to 1900 time steps. Such closure will alter the travel distance between nearby nodes resulting in modified matrix $\mathbf{Q}^{(t)}$. 
Besides the evaluation on the entire dataset, we also highlight the results of node 8 and its nearby roads 7 and 9, whose surroundings changed the most. 

\subsubsection{Baselines}
We compare our model with the following baselines on the two real-world datasets. The default settings of these methods are used.
\begin{itemize}
	\item HA: Historical average method. We use the moving average with window size 12 for the forecasting.
	\item ARIMA$_{kal}$ \cite{williams2003modeling}: Auto-Regressive integrated moving average model with Kalman filter for time series analysis. 
    \item FC-LSTM \cite{Sutskever2014Deep}: Recurrent neural network with fully connected LSTM hidden units.
	\item DCRNN \cite{li2017diffusion} combines diffusion convolution and gated recurrent units in an encoder-decoder manner. 
	\item STGCN \cite{yu2017spatio} employs graph convolution for spatial patterns and 1D convolution for temporal features.  
	\item Graph WaveNet \cite{wu2019graph} uses node embeddings to learn a dependency matrix, and the dilated 1D convolution for prediction. 
	\item GMAN \cite{zheng2020gman}: Graph multi-attention network equipped with spatial, temporal and transform attention.
	\item MTGCN \cite{wu2020connecting}: Multivariate time series forecasting with
	graph neural networks. 
\end{itemize}

\subsubsection{Experiment Settings}
We implemented our model via the PyTorch framework and on the following hardware platform: (CPU: Intel(R) Xeon(R) Gold 6128 CPU @ 3.40GHz, GPU: NVIDIA Quadro RTX 8000). For the STNN model, we set $k=2$, i.e.,  two ST-Modules are stacked, which is enough to produce satisfying results. The number of output channels for two ST-Modules are 32 and 64, respectively. The convolutional kernel size is set to be $f=3$. The parameter used in the LeakyReLU is 0.2. In the experiments, datasets are split as 70\% training, 20\% validation and 10\% testing in chronological order. The length of input and output time sequence are both 12. In the training phase, the batch size is 80 , and the number of epochs is 50, which takes about 5 hours to train the model on METR-LA dataset. Adam optimizer is adopted with learning rate 1$\times$ 0.001 to update model parameters towards the minimized L1 loss between the predicted value and the grounded truth. Dropout is enabled with rate 0.3. We apply the commonly used metrics: mean absolute error (MAE), root mean squared error (RMSE), and mean absolute percentage error (MAPE) to evaluate the mean prediction accuracy of future $T_r$ time steps. We report the average performance of five runs. Random seed of PyTorch is set to be 1 for reproducibility purpose. 

We perform a grid search for several important hyper-parameters to determine the appropriate values regarding complexity and performance. The first hyper-parameter is the number of nodes in each local-spacetime. The second is the proportion of data used for training, ranges from 0.05 to 1. The third is the number of spacetime modules employed, ranges from 1 to 4. Results show that the model performance improves drastically with the increasing number of neighbors in a local-spacetime and reach the plateau after $\epsilon=15$. Similar pattern shows for the ratio of training set and the best-tradeoff point is 0.2. In other words, in 70\% training data, we random select $0.2\times 70\%$ for the final training. Last, model is not very sensitive to the number of ST modules after stack two modules. 

\begin{table*}[t]
	\centering
	\small
	\caption{Overall performance of short-term (15 mins), mid-term (30 mins) and long-term (60 mins) traffic forecasting. 
	}
	\begin{tabular}{l|l|l|ccccccccc}
		\toprule
		\multirow{2}{*}{Training Data} 
		&\multirow{2}{*}{Test Data}
		&\multirow{2}{*}{Models} 
		& \multicolumn{3}{c}{15 mins} 
		& \multicolumn{3}{c}{30 mins} 
		& \multicolumn{3}{c}{60 mins} \\ \cline{4-12} 
		&  & & MAE & RMSE & \multicolumn{1}{l|}{MAPE} & MAE & RMSE & \multicolumn{1}{l|}{MAPE} & MAE & RMSE & MAPE \\ \midrule
		
		\multirow{8}{*}{METR-LA}& \multirow{8}{*}{METR-LA}  
		& HA 
		& 4.16 & 7.80 & \multicolumn{1}{c|}{13.02\%} 
		& 4.16 & 7.80 & \multicolumn{1}{c|}{13.02\%} 
		& 4.16 & 7.80 & 13.02\% \\
		
		& & ARMIA 
		& 3.60 & 7.18 & \multicolumn{1}{c|}{8.95\%} 
		& 4.37 & 8.72 & \multicolumn{1}{c|}{10.42\%} 
		& 5.56 & 10.01 & 13.50\% \\
		
		& & FC-LSTM 
		& 3.02 & 5.27 & \multicolumn{1}{c|}{8.92\%} 
		& 3.29 & 6.78 & \multicolumn{1}{c|}{9.10\%} 
		& 3.71 & 7.32 & 9.54\% \\
		
		& & DCRNN 
		& 2.52 & 4.96 & \multicolumn{1}{c|}{6.81\%} 
		& 2.71 & 5.57 & \multicolumn{1}{c|}{7.85\%} 
		& 3.03 & 6.15 & 8.84\% \\
		
		& & STGCN 
		& 2.76 & 5.01 & \multicolumn{1}{c|}{6.90\%} 
		& 2.94 & 6.17 & \multicolumn{1}{c|}{8.26\%} 
		& 3.85 & 7.50 & 9.73\% \\
		
		& & Graph WaveNet 
		& 2.36 & 4.54 & \multicolumn{1}{c|}{6.04\%} 
		& 2.59 & 5.25 & \multicolumn{1}{c|}{7.05\%} 
		& 2.98 & 6.07 & 8.21\% \\
		
		& & GMAN 
		& 2.39 & 4.63 & \multicolumn{1}{c|}{5.95\%} 
		& 2.58 & 5.31 & \multicolumn{1}{c|}{6.93\%} 
		& \textbf{2.94} & \textbf{6.02} & \textbf{8.16\%} \\
		
		& & MTGCN 
		& 2.41 & 4.58 & \multicolumn{1}{c|}{6.01\%} 
		& 2.59 & \textbf{5.23} & \multicolumn{1}{c|}{6.98\%}
		& 2.96 & 6.04 & 8.18\% \\
		
		& & \textbf{STNN (ours)} 
		& \textbf{2.27} & \textbf{4.46} & \multicolumn{1}{c|}{\textbf{5.80\%}} 
		& \textbf{2.56} & 5.29 & \multicolumn{1}{c|}{\textbf{6.84\%}} 
		& 3.01 & 6.23 & 8.50\% \\
		
		\hline
		\multirow{8}{*}{PeMS-Bay} & \multirow{8}{*}{PeMS-Bay}
		& HA 
		& 2.88 & 5.59 & \multicolumn{1}{c|}{6.85\%} 
		& 2.88 & 5.59 & \multicolumn{1}{c|}{6.85\%}
		& 2.88 & 5.59 & 6.85\% \\
		
		& & ARMIA 
		& 1.55 & 3.01 & \multicolumn{1}{c|}{3.12\%} 
		& 1.98 & 4.12  & \multicolumn{1}{c|}{4.76\%}
		& 2.71 & 5.34 & 6.62\% \\
		
		& & FC-LSTM 
		& 1.89 & 3.88 & \multicolumn{1}{c|}{4.40\%} 
		& 2.01 & 4.19 & \multicolumn{1}{c|}{4.75\%}
		& 2.16 & 4.38 & 5.24\% \\
		
		& & DCRNN 
		& 1.28 & 2.63 & \multicolumn{1}{c|}{2.57\%} 
		& 1.51 & 3.49 & \multicolumn{1}{c|}{3.40\%} 
		& 1.78 & 4.19 & 4.28\% \\
		
		& & STGCN 
		& 1.27 & 2.65 & \multicolumn{1}{c|}{2.60\%} 
		& 1.58 & 3.88 & \multicolumn{1}{c|}{3.70\%} 
		& 2.20 & 4.63 & 4.87\% \\
		
		& & Graph WaveNet 
		& 1.22 & 2.45 & \multicolumn{1}{c|}{2.50\%} 
		& \textbf{1.47} & 3.31 & \multicolumn{1}{c|}{3.21\%} 
		& 1.68 & 3.56 & 3.88\% \\
		
		& & GMAN 
		& 1.26 & 2.50 & \multicolumn{1}{c|}{2.58\%} 
		& 1.47 & 3.32 & \multicolumn{1}{c|}{\textbf{3.18\%}} 
		& \textbf{1.65} & \textbf{3.53} & \textbf{3.61\%} \\
		
		& & MTGCN 
		& 1.24 & 2.46 & \multicolumn{1}{c|}{2.53\%} 
		& 1.50 & 3.34 & \multicolumn{1}{c|}{3.30\%}
		& 1.69 & 3.65 & 3.90\% \\ 
		
		& & \textbf{STNN (ours)} 
		& \textbf{1.20} & \textbf{2.41} & \multicolumn{1}{c|}{\textbf{2.50\%}} 
		& 1.49 & \textbf{3.23} & \multicolumn{1}{c|}{3.26\%}
		& 1.86 & 4.17 & 4.30\% \\ 
		
		\hline
		\hline
		\multirow{1}{*}{PeMS-Bay} & \multirow{1}{*}{METR-LA}
		&  \multirow{4}{*}{\textbf{STNN (ours)}}
		& 2.60 & 5.02 & \multicolumn{1}{c|}{6.75\%} 
		& 3.01 & 5.94 & \multicolumn{1}{c|}{8.37\%} 
		& 3.68 & 7.33 & 11.13\% \\
		   
	    \cline{1-2}\cline{4-12}
		\multirow{1}{*}{METR-LA} & \multirow{1}{*}{PeMS-Bay}
		& 
		& 1.36 & 2.73 & \multicolumn{1}{c|}{2.92\%} 
		& 1.68 & 3.58 & \multicolumn{1}{c|}{3.73\%}
		& 2.15 & 4.72 & 5.01\% \\ 
		
		\cline{1-2}\cline{4-12}
		\multirow{2}{*}{Combined} & METR-LA
		&  
		& {2.30} & {4.50} & \multicolumn{1}{c|}{{5.83\%}} 
		& {2.60} & {5.29} & \multicolumn{1}{c|}{{6.90\%}}
		& {3.07} & {6.37} & {8.80\%} \\
		
		\cline{2-2}\cline{4-12}
		 & PeMS-Bay
		&  
		& {1.21} & {2.43} & \multicolumn{1}{c|}{{2.51\%}} 
		& {1.49} & {3.22} & \multicolumn{1}{c|}{{3.26\%}}
		& {1.87} & {4.17} & {4.40\%} \\
		\bottomrule
	\end{tabular}

	\label{tab:results_comparison}
\end{table*}

\subsection{Evaluation on Real-World Data}\label{sec:realworld}
One salient feature of this work is that a trained STNN model can be used directly on unseen traffic networks. None of the existing state-of-the-art methods can do this. Current models train and predict on the same dataset, and the trained model cannot be applied to other networks with different sizes. Thus, we evaluate the performance of STNN in three settings: (1) train and test on the same network; (2) train and test on different networks; (3) train on multiple networks and test on each network.

\subsubsection{Train and Test on the Same Network}
This experiment follows the conventional settings where we train and test the STNN model on the same network.  
We train two STNN models separately on the two real-world networks, namely $\text{STNN}_1$ for METR-LA dataset and $\text{STNN}_2$ for PeMS-bay dataset, respectively. The prediction accuracy on test sets are shown in the first two rows of Table~\ref{tab:results_comparison}. STNN surpasses the baselines in both datasets. For short period predictions like the METR-LA 15 mins case, $\text{STNN}_1$ achieves 2.27 MAE and 4.46 RMSE. Compared with the baselines, where the best MAE and RMSE are 2.36 and 4.54, respectively, $\text{STNN}_1$ produces up to 4\% performance improvement. For mid-term predictions (30 mins), STNN can reach the SOTA performance. For long period predictions like 60 mins case, our model is slightly worse than the baselines like GMAN which designed specifically for long-term traffic prediction.

\subsubsection{Train and Test on Different Networks} Since STNN is designed to discover universal traffic patterns. An interesting question is to apply a trained model on an unseen dataset. This is a huge challenge for methods that rely on the graph adjacency/Laplacian matrix because these matrices can be dramatically different in various networks. 
The proposed STNN concentrates on the local spatial and temporal context such that the learned model can be easily used on unseen datasets. In previous experiments, we have trained the $\text{STNN}_1$ on METR-LA and $\text{STNN}_2$ on PeMS-bay. Without any fine-tuning, we take $\text{STNN}_1$ to make predictions on PeMS-bay data and the performance is comparative to state-of-the-art methods trained directly on PeMS-bay. Similar results for $\text{STNN}_2$ on METR-LA. The details are shown in the middle two rows of Table~\ref{tab:results_comparison}. The MAE of $\text{STNN}_1$ on PeMS-bay 15 mins prediction period is 1.36, that is comparable to existing methods which train and test on the same dataset. The MAE of $\text{STNN}_2$ on METR-LA 15 mins prediction period is 2.60 while the MAEs of baselines range from 2.39 to 2.76.

\subsubsection{Train on Multiple Networks}
Finally, we trained a single model on the combined dataset by mixing the training examples from PeMS-Bay and METR-LA. The learned STNN model captures the general traffic patterns from both datasets. The results are shown in the last two rows of Table~\ref{tab:results_comparison}. The performance of the uniform STNN is very close to STNNs that train and test on the same network, and outperforms baselines on both datasets of short-term prediction. 

\vspace{-0.2cm}
\subsection{Evaluation on Simulated Dynamic Network}\label{sec:simulated}
Table~\ref{tab:results_comparisons2} shows the traffic volume prediction error for the next three time steps of nodes 7, 8, 9, and the average error on the whole network. We only compare with HA, ARMIA and FC-LSTM because the other methods cannot be applied on networks with dynamic topology. In general, STNN outperforms the baselines by large margin. The overall MAE and RMSE of STNN are 5.31 and 19.83 for the entire dataset, which is 42\% better than baselines. Moreover, for a specific location like node 7, the MAE and RMSE of the baseline are 9.75 and 19.39, respectively. STNN achieves 5.27 MAE and 10.31 RMSE, leading to the 46\% and 47\% improvement against the baseline. For the road segmentation monitored by sensor 8, the traffic volume is always zero during the closure period. It is not very sensible to predict the traffic flow for such time steps. Thus, we exclude them during evaluation. Similar to other nodes, STNN still outperforms baselines with a 35\% better MAE. It is worthy to note that the prediction error on node 7 is larger than node 9, mainly because node 7 is one of the traffic sources where new vehicles keep entering the traffic network continuously and randomly makes it hard to predict. 

\begin{table}
    \footnotesize
	\centering
	\caption{Performance of traffic volume prediction on the simulated dynamic network. We highlight sensor 7,8,9 as they are most affected by the changing network topology.}
	\label{tab:results_comparisons2}
	\begin{tabular}{llccc}
		\hline
		Data & Models & MAE & RMSE & MAPE\\

		\hline \multirow{4}{*}{Node 7} 
		& HA & 26.69  &  64.57  &  12.53 \%\\
		& ARMIA & 12.08  & 29.16  &  9.84 \%\\
		& FC-LSTM& 9.75  & 19.39  &  7.74 \% \\
		& \textbf{STNN (ours)} 
		& \textbf{5.27} & \textbf{10.31} & \textbf{3.97 \%} \\

		\hline
		\multirow{4}{*}{Node 8} 
		& HA & 20.75 & 43.82   &  28.79 \%\\
		& ARMIA & 17.37  & 35.22 &  23.54 \%\\
		& FC-LSTM& 15.19  &  30.46 & 18.69  \% \\
		& \textbf{STNN (ours)} 
		& \textbf{9.78 } & \textbf{26.31 } & \textbf{9.63 \%} \\
		
		\hline
		\multirow{4}{*}{Node 9} 
		& HA &  11.75 &  29.63  &  10.55 \%\\
		& ARMIA & 6.35  &  11.35 &  6.10 \%\\
		& FC-LSTM& 5.23  & 9.61  & 5.19  \% \\
		& \textbf{STNN (ours)} 
		& \textbf{2.65} & \textbf{6.14 } & \textbf{ 2.88 \%} \\
		
		\hline
		\multirow{4}{*}{Overall} 
		& HA & 15.53 & 42.19 &  10.01 \% \\
		& ARMIA & 11.02 & 31.75 & 7.81 \% \\
		& FC-LSTM & 9.17 & 25.65 &  5.89\% \\
		& \textbf{STNN (ours)} & \textbf{5.31} & \textbf{19.83} & \textbf{3.79\%} \\
		\hline
	\end{tabular}
\end{table}

\subsection{Case Study}
\label{sec:case_study}
\begin{figure}[htb!]
\centering
\begin{subfigure}[b]{0.45\textwidth}
\centering
  \includegraphics[width=0.6\linewidth]{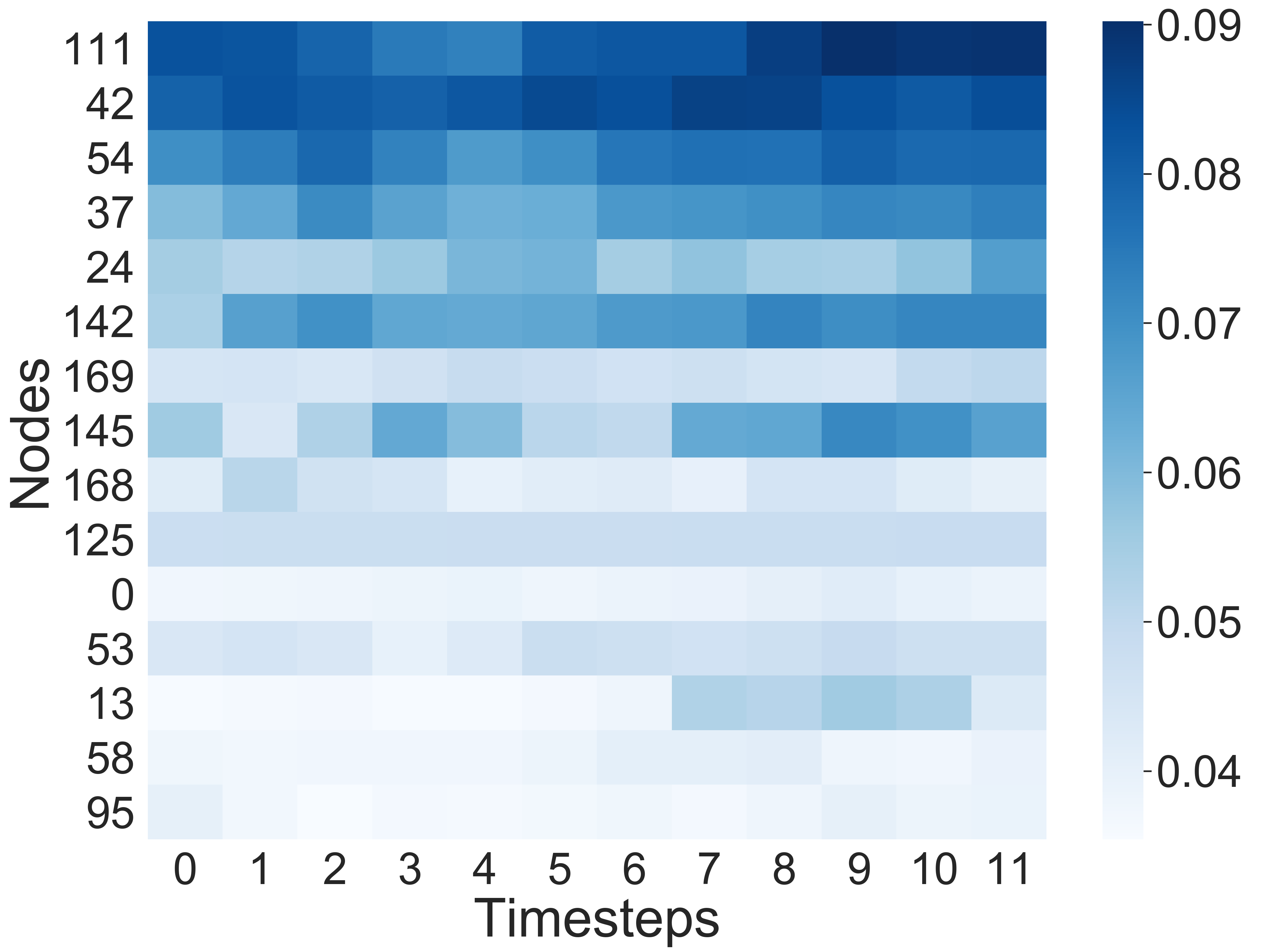}
  \caption{Impact of traffic events for the prediction of future traffic flow at node 111. Nodes are sorted by travel distance with 111 on road network.}
  \label{fig:case_study1} 
\end{subfigure}

\begin{subfigure}[b]{0.45\textwidth}
\centering
  \includegraphics[width=0.7\linewidth]{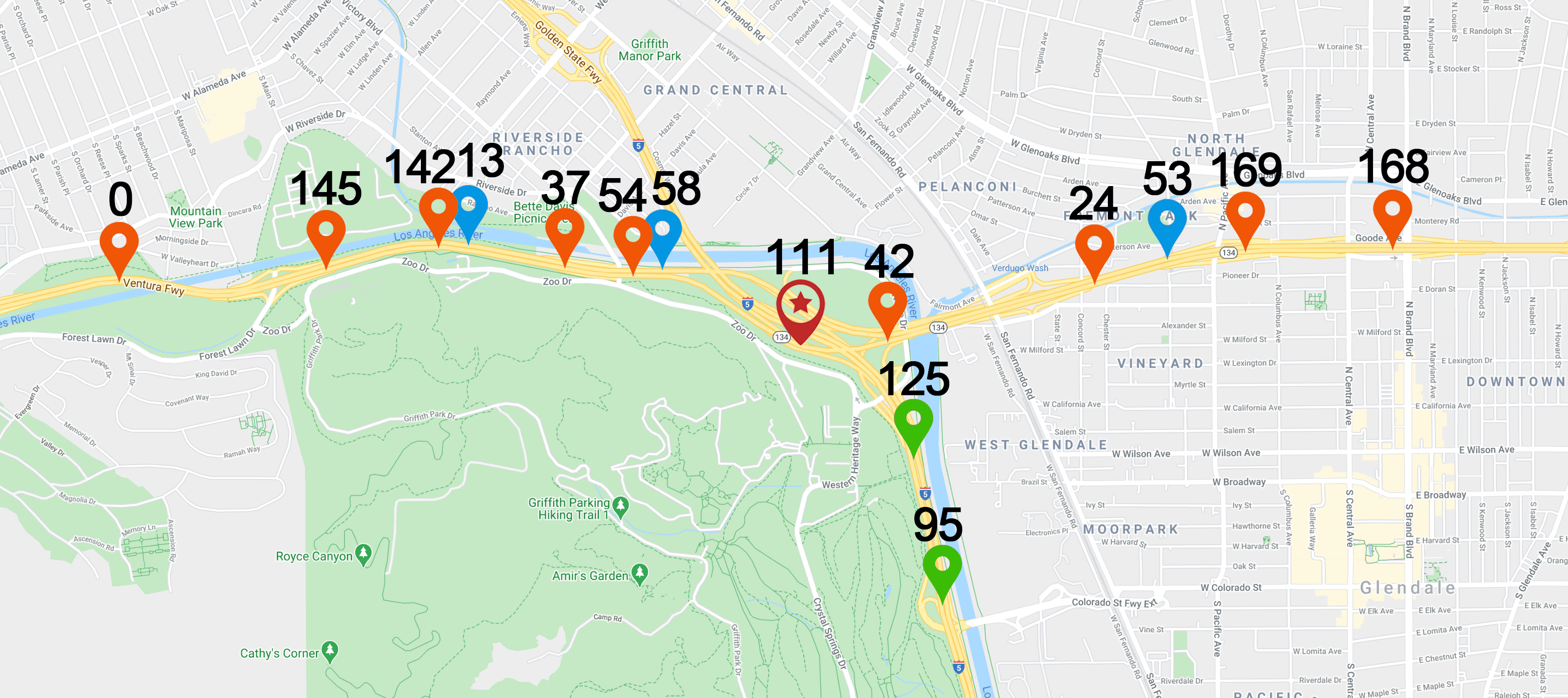}
  \caption{Spatial context of node 111. Red markers represent roads from west to east, blue markers represent roads from east to west, and green markers stand for north to south.}
  \label{fig:case_study2}
\end{subfigure}
\caption{Case study}
\label{fig:case_study}
\end{figure}

To develop a better insight of our model, we present a real-world case study. The aim is to reveal the spacetime intervals that are extracted by our model.  Figure~\ref{fig:case_study} illustrates a small part of the METR-LA dataset that contains a number of nodes around the Los Angeles Zoo. These nodes are used for the prediction of a target node, as indicated by node 111. We set the prediction period to 15 minutes (3 time steps). The color scale in Figure~\ref{fig:case_study1}, which is transformed from the learned attention map, indicates the different influence (i.e., spacetime interval) made by those surrounding traffic events have towards the target node 111. The vertical axis represents the spatial dimension where nodes are listed in ascending order by their distance with the target node 111. The horizontal axis represents the temporal dimension where 0--11 time steps data are used for predicting the next 3 time steps of node 111. Figures \ref{fig:case_study2} displays the spatial context of all nodes involved. From these two figures, we can observe that the most significant nodes for prediction are 111, 42, 54, 37, 142, 145. Among them, 54, 37, 142, 145 are the nearest upstream neighbors; and 42 is the immediate downstream neighbor. It is remarkable that downstream locations could be useful for traffic forecast, while the correlation fades out rapidly as distance increases as for node 24. Besides the spatial context, the temporal dimension also plays an important role. The time steps of node 111 that are key to the prediction are the immediate last four while this is not true for other nodes. Despite close proximity of nodes 125 and 58 to 111, they demonstrate a smaller influence as 58 is on the opposite direction of the same road while 125 is on another road. This case study demonstrates that our model can extract meaningful correlations between the traffic events along both the spatial and temporal dimensions.

\subsection{Ablation Study}\label{sec:ablation}
We perform an ablation study to validate the impact of each proposed component to the model performance. In each experiment, a block/layer is removed and the rest remain unchanged. Particularly, six types of experiments are conducted: full STNN, STNN without ST-Attn block, STNN without ST-Conv block, STNN without spatial convolution layer, STNN without temporal layer, and STNN without spacetime convolution layer. We report the performance of the above models for short-term (15 mins) forecasting in Table~\ref{tab:ablation_study}.

\begin{table}[!htbp]
\small
\caption{Ablation study}
\label{tab:ablation_study}
\begin{tabularx}{\linewidth}{lXXXXXX}
\toprule
& \multicolumn{3}{c}{METR-LA} & \multicolumn{3}{c}{PeMS-Bay} \\
& MAE & RMSE & MAPE & MAE & RMSE & MAPE \\
\midrule
\textbf{Full STNN} & \textbf{2.27} & \textbf{4.46} & \textbf{5.80} & \textbf{1.20} & \textbf{2.41} & \textbf{2.53} \\
No ST-Attn block & 2.35 & 4.56 & 5.85 & 1.23 & 2.43 & 2.55 \\
No ST-Conv block & 2.41 & 4.69 & 5.97 & 1.30 & 2.58 & 2.71 \\
No spatial conv & 2.31 & 4.51 & 5.74 & 1.22 & 2.42 & 2.50 \\
No temporal conv & 2.33 & 4.58 & 5.90 & 1.22 & 2.47 & 2.55 \\
No spacetime conv & 2.34 & 4.59 & 5.98 & 1.23 & 2.49 & 2.58\\
\bottomrule
\end{tabularx}
\end{table}

The effect of proposed spacetime attention block is evident since it highlights the contributive traffic events. The spacetime convolution block improves the prediction accuracy significantly as it aggregated the traffic events in three perspectives. Moreover, the spacetime convolutional layer appears to be more important than the other two convolutional layers.

\subsection{Complexity Analysis}\label{sec:complexity}
We analyze the complexity of the proposed model to justify its scalability by showing the computation of STNN is independent of the road network size. In addition, the total trainable parameters in STNN is 308,786, which is smaller than STGCN (454,358), MTGCN (405,452), and other state-of-the-art methods.

The time complexity of the spacetime convolutional block is $O(4\alpha T d_i d_o)$ where $\alpha$ is the number of sensors in local-spacetime, $T$ is the length of input time steps, $d_i$ and $d_o$ are channels of input feature maps and output feature maps. Additionally, four different convolutional layers (spacetime, spatial, temporal, 1*1) were utilized in the spacetime convolutional block. The spacetime attention block incurs $O(c_ic_o(\alpha^2 T^2 + \alpha T))$ time complexity where $c_i$ is the channels of input data and $c_o$ is the channels of output data. $\alpha$ and $T$ still represents the size of the spacetime manifold. In short, the computational complexity of the proposed STNN depends on the local-spacetime size, input sequence length, and channels. However, it is independent of the entire network.

Last, the local-spacetime construction is part of the proposed spacetime interval learning framework which is not included in the forward pass and backpropagation of STNN. The complexity of local-spacetime construction equals $O(NT)$ where $N$ is the number of sensors in the network. 

\section{Conclusion}
In this paper, we propose a novel spacetime interval learning framework and spacetime neural network (STNN) for accurate traffic prediction. Our approach captures the intrinsic principles of traffic flow by learning the spacetime intervals between traffic events. The model works on the local-spacetime extracted for target nodes and thus can handle dynamic network topology. Experiments on two real-world datasets and a simulated dynamic network show that the proposed framework significantly outperforms state-of-the-art methods. This confirms the effectiveness of capturing spatio-temporal correlations directly.
In the future, we will further explore the possibility of capturing the graph dynamics from the input data when the underlying graph structure is unknown. Another promising direction is to embed the local-spacetime construction into an end-to-end deep learning framework, making the parameters used in local-spacetime construction learnable. Last, we would like to investigate the spacetime interval learning paradigm in other spatio-temporal data modeling scenarios.

\bibliographystyle{IEEEtran}
\bibliography{references}

\end{document}